\documentclass[letterpaper, 10 pt, conference]{ieeeconf}  

\IEEEoverridecommandlockouts                              

\usepackage{graphics} 
\usepackage{graphicx}
\usepackage{epsfig} 
\usepackage{amsmath} 
\usepackage{amssymb}  
\usepackage{dsfont}
\usepackage{float}
\usepackage{comment}
\let\labelindent\relax
\usepackage{enumitem}
\usepackage{hyperref}
\usepackage{multirow}
\usepackage{microtype}

\usepackage{color}
\definecolor{orange}{RGB}{255,229,204} 
\definecolor{lred}{RGB}{255,204,204} 
\definecolor{lgreen}{RGB}{200,240,200} 
\definecolor{lblue}{RGB}{204,229,255} 

\definecolor{darkbrown}{rgb}{0.4, 0.26, 0.13}
\definecolor{amethyst}{rgb}{0.6, 0.4, 0.8}
\definecolor{blue-violet}{rgb}{0.54, 0.17, 0.89}
\definecolor{caputmortuum}{rgb}{0.35, 0.15, 0.13}
\definecolor{darkviolet}{rgb}{0.58, 0.0, 0.83}
\definecolor{lavender}{rgb}{0.9, 0.9, 0.98}
\definecolor{indigo}{rgb}{0.29, 0.0, 0.51}

\newcommand{\paolo}[1]{{\color{darkbrown}{#1}}} 

\newcommand{\ours}{D3G}

\usepackage{pifont}

\usepackage{comment}

\title{\LARGE \bf
A Modern Take on \\ Visual Relationship Reasoning for Grasp Planning
}

\author{ Paolo Rabino$^{1}$ and Tatiana Tommasi$^{1}$
\thanks{$^{1}$Paolo Rabino and Tatiana Tommasi are with the Department of Control and Computer Engineering, at Politecnico di Torino, Italy
        {\tt\small \{paolo.rabino, tatiana.tommasi\}@polito.it} 
    }
}

\begin{document}

\maketitle
\thispagestyle{empty}
\pagestyle{empty}


\begin{abstract}
        Interacting with real-world cluttered scenes poses several challenges to robotic agents that need to understand complex spatial dependencies among the observed objects to determine optimal pick sequences or efficient object retrieval strategies.
Existing solutions typically manage simplified scenarios and focus on predicting pairwise object relationships following an initial object detection phase, but often overlook the global context or struggle with handling redundant and missing object relations.
In this work, we present a modern take on visual relational reasoning for grasp planning. We introduce D3GD, a novel testbed that includes bin picking scenes with up to 35 objects from 97 distinct categories. Additionally, we propose D3G, a new end-to-end transformer-based dependency graph generation model that simultaneously detects objects and produces an adjacency matrix representing their spatial relationships.
Recognizing the limitations of standard metrics, we employ the Average Precision of Relationships for the first time to evaluate model performance, conducting an extensive experimental benchmark. 
The obtained results establish our approach as the new state-of-the-art for this task, 
laying the foundation for future research in robotic manipulation. 
We publicly release the code and dataset at \url{https://paolotron.github.io/d3g.github.io}.
\end{abstract}
\section{Introduction}

Every kitchen has a cluttered drawer where that specific tool you urgently 
need to complete a recipe is buried beneath various gadgets and stray items. Retrieving it requires carefully planning a sequence of grasps. Could robot assistants help with this task?
As they progressively integrate into our daily lives, we would like to delegate increasingly complex procedures composed of multiple key steps to artificial agents.
This requires comprehensive perception, understanding of highly unstructured environments and advanced planning skills. 
Kitchen drawers as well as industrial containers are just two of many possible examples of multi-object-stacking scenes requiring appropriate grasping sequences to prevent potential damages.

Previous research has framed this problem as \emph{visual manipulation relationship reasoning} \cite{vrmn, gvrmn_neuro, GruVMRN}, highlighting as a primary challenge the inference of relationship among objects, which can be categorized as parent, child, or none. In this context, objects are nodes in a dependency graph, where parent nodes should be grasped after child nodes and this information is encoded in the edges.  
Knowing the complete dependency graph enables the calculation of the shortest path to any object and facilitates planning the most efficient picking sequence to access it. 

\begin{figure}
    \centering
\includegraphics[width=0.80\linewidth]{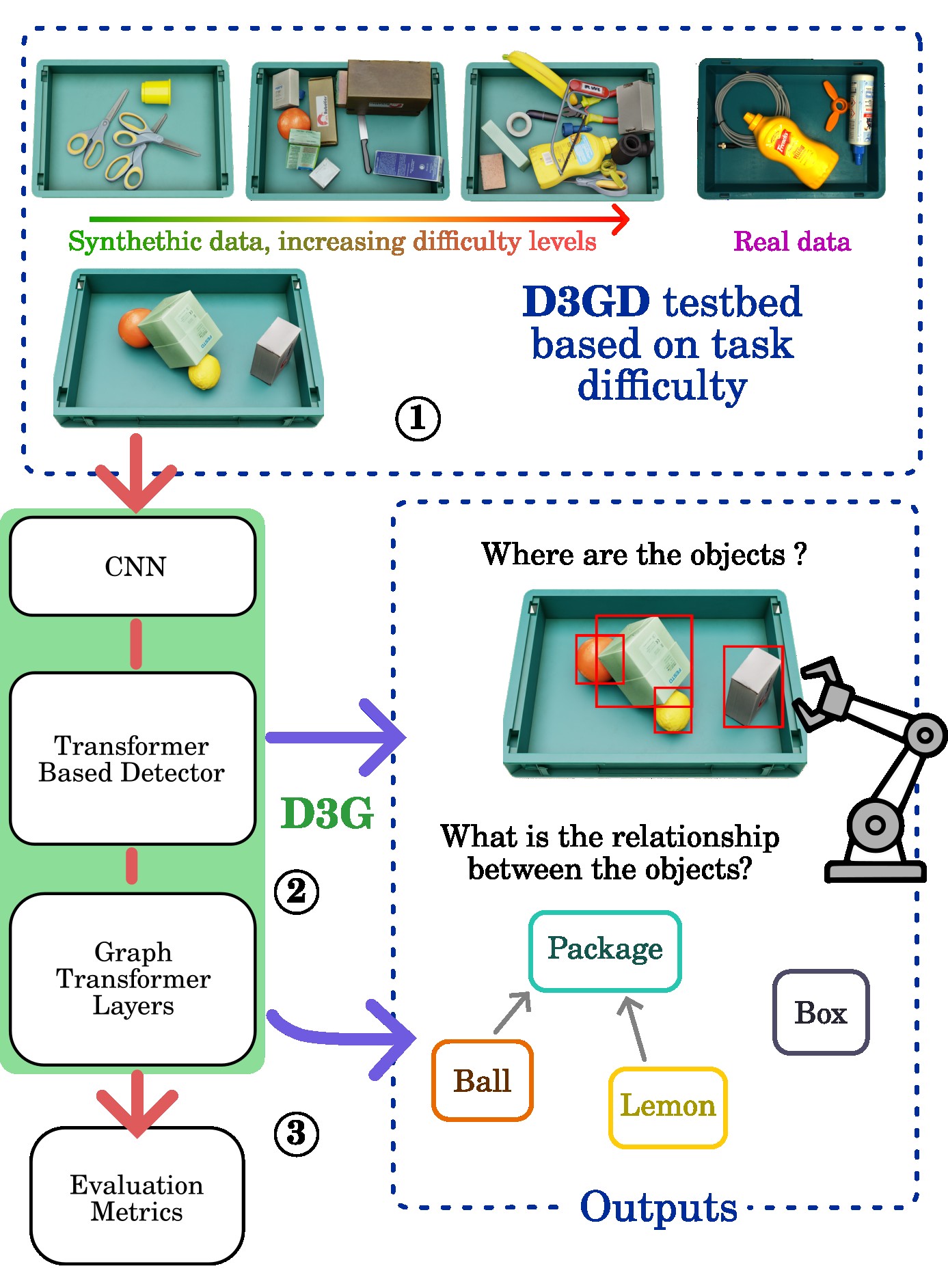} \vspace{-3mm}
    \caption{Manipulation relationship graphs provide essential information to plan grasping in complex scenes as those encountered in industrial bin picking. Our work introduces a new testbed for this task \ding{192}.  
    Moreover, we design a novel approach 
    \ding{193} that considers both object detection and edge relationship prediction in an end-to-end process.
    Finally, we adopt a tailored 
    evaluation metric \ding{194} to assess the method performance across different levels of difficulties and challenges. 
}
    \label{fig:teaser} \vspace{-6mm}
\end{figure}

Existing algorithms address this task by decoupling object localization from relationship reasoning. They first identify the objects through a detector and then encode their relationship via a handcrafted post-hoc procedure known as object pair pooling \cite{vrmn, gvrmn_neuro, GruVMRN}. This technique crops the learned feature map based on the union area of two objects to encode their relationship. However, in highly cluttered scenes, the cropping region may disproportionately represent one object over the other or include irrelevant objects. 
As a result, the generated representation is ambiguous and leads to inconsistencies in the relationship models, particularly in addressing issues such as redundant or missing edges \cite{E-VRMN, vrmn, gvrmn_neuro}.

While these shortcomings may not be immediately apparent in simplified scenarios involving a small number of objects (e.g. 5 objects as in \cite{vrmn}) or moderate setups (e.g. 19 objects in \cite{Regrad}) on a tabletop, they become increasingly evident as the task complexity scales. A bin filled with a large number of objects provides a compelling test case, as the bin's boundaries facilitate object stacking, resulting in dense and intertwined layouts that correspond to highly complex dependency graphs.
Moreover, the evaluation metrics commonly used to assess visual manipulation relationship reasoning methods are derived from detection literature and are influenced by an implicit object scoring threshold. This threshold is dataset-specific, challenging to calibrate, and may hinder the generalization of the evaluation process.

With our work we propose a modern take on 
visual relationship reasoning for grasp planning 
and advance the field with the following contributions (see Fig. \ref{fig:teaser}): 

\smallskip
\textbf{(1) We introduce a new testbed} based on the MetaGraspNetV2 dataset \cite{metagraspnetv2} that presents synthetic and real bin picking scenes annotated with dependency graphs. 
We manually organized the scenes   
in subsets of growing complexity by focusing on the number of nodes, connected subgraphs (clutter width), and shortest paths from leaf to root (clutter depth). 
We name it D3GD:  \emph{Dependency Graph Generation for Grasp planning Dataset}, 
and contains scenes with up to 35 objects from 97 different categories.

\smallskip
\textbf{(2) We design a new end-to-end dependency graph generation model.} We dub it D3G. 
Its objective is to jointly detect the objects and predict the adjacency matrix describing their spatial relationship. 
We formalize the task as an object set prediction problem over the entire image: leveraging global context, the model chooses where each object query should focus and a graph transformer ensures that relationship features attend to the relevant object queries. 
This enables simultaneous learning of object locations, appearances, and representation for their connecting edges.

\smallskip
\textbf{(3) We assess the entire pipeline with a tailored metric.} 
We use for the first time the \emph{Average Precision of Relationships} \cite{kuznetsova2020open} to evaluate the generated dependency graph. 
This metric calculates precision-recall trade-offs over predicted triplets defined by two objects and their relation. It provides a threshold-free evaluation score that extends beyond conventional object detection metrics.
With a thorough experimental analysis we show how our approach outperforms several state-of-the art competitors in challenging experimental settings. Overall our work presents a novel benchmark suite that will be open-sourced to the community paving the way for future application-oriented research in robotic manipulation.
\section{Related Works}
In this section we provide a brief overview of previous publications in research areas related to our work, starting from the basic components of object detection and graph learning. We also review existing literature on object relationship recognition and reasoning, considering applications to scene understanding and robotic manipulation. \\
\textbf{Object Detection} is an extensively studied task that aims to localize objects in a scene and recognize their category label. 
Detection approaches evolved from feature matching and part modeling \cite{Felzenszwalb2010ObjectDW}, through two stages  \cite{faster-rcnn,mask-rcnn} and one stage deep methods \paolo{\cite{Liu2015SSDSS}}, getting more recently to transformer-based techniques. The latter allows for global computation over a whole image and set prediction, leveraging attention between multiple objects as in the case of DETR \cite{detr}.
Deformable attention has proven to enhance such approaches \cite{zhu2020deformable}. 
Further enhancements have been obtained through query-denoising \cite{li2022dn} and dynamic anchors \cite{liu2022dabdetr, zhao2024detrs}.\\
\textbf{Graph Learning} effectively encodes complex interactions in structured data.
Message passing based networks have been designed to leverage the sparsity and connectivity of graph structures \cite{Kipf2016SemiSupervisedCW}, efficiently propagating information and capturing dependencies between nodes \cite{wu2020comprehensive}. 
With the advent of self-attention based architectures \cite{attallneed}, several efforts have been made to merge the graph neural network paradigm and the transformer one, obtaining competitive performance on graph benchmarks \cite{yun2019graph, dwivedi2021generalization, rampavsek2022recipe, Xia2024OpenGraphTO}. \\
\textbf{Relationship reasoning} encompasses tasks where it is important to model geometric (e.g. spatial positioning) or semantic relations (e.g. subject-object interactions) from data.
Large datasets such as Visual Genome \cite{Krishna2016VisualGC} and Open Images  \cite{kuznetsova2020open} paved the way for perception studies mainly dedicated to image captioning, with first solutions building on two-stage object detectors \cite{lu2016visual}, and the latest ones based on transformers \cite{im2024egtr} \cite{scenegraphVIT2024}.
Several research works in robotics have been dedicated to dependency graph generation that aims at mapping the relationships and sequential constraints among multiple objects in highly cluttered scenes to support decision-making processes and motion planning \cite{Zhu2020HierarchicalPF, gu2024conceptgraphs}.

Such a graph is used as an intermediate result \cite{StackPlanning, Huang2022PlanningFM} or an auxiliary objective for enhancing grasp detection models \cite{Zhang2018AMC, Tchuiev2022}. In \cite{Xiong2021AGI} dependency graphs 
 are exploited to decide grasping order for efficient bin emptying, \cite{motoda2021bimanual} uses them to predict the likelihood of object stacks collapse, and \cite{li2024broadcasting} leverages them to plan the optimal path to retrieve specific objects in dense clutter. \\
The first real-world visual manipulation relationship dataset (VMRD) was introduced in \cite{vrmn}. It is a collection of 5185 hand-annotated images depicting various configurations of objects in cluttered scenes each containing up to 5 object instances sampled from 31 categories. While valid, this dataset is limited in dimension and complexity. The authors also proposed a deep model with distinct object detection and relationship prediction stages. The same high-level solution was then adopted by following approaches that used fully connected conditional random fields \cite{E-VRMN} and graph neural networks \cite{gvrmn_neuro, GruVMRN}. These works included analyses on a larger relational grasp dataset named REGRAD \cite{Regrad} automatically collected in a virtual environment\footnote{Reproducible results on a dataset require knowledge of the pre-processing pipeline and access to baseline code and checkpoints. Unfortunately, most prior work in visual manipulation relationship reasoning lacks these resources. Specifically, in REGRAD, the detailed analysis uncovered issues such as missing boxes, one-pixel masks, and inconsistent object classes. The lack of public code further impedes result replication. We overcome these limitations with our benchmark, re-implementing the competitor methods and releasing the code for fair comparison.}. It comprises images annotated with manipulation relationships between objects in dense clutter and includes multi-view perspectives \cite{wang2023mmrdn}. 
Nonetheless, existing testbeds tend to focus on relatively simple scenarios and often lack comprehensive information on the experimental settings, making previous results difficult to reproduce.
Lately with the release of the NVIDIA Isaac simulator \cite{makoviychuk2021isaac} a new generation of highly realistic and accurately annotated datasets have been published \cite{metagraspnetV1, metagraspnetv2, Wang2022DexGraspNetAL}, which provide the opportunity to scale up relational reasoning related tasks for robotics applications.
\section{Dataset}

\begin{table}
\centering
    \caption{Difficulty levels 
    of our D3GD dataset. Each row describes the concurrent conditions that define a level set. 
    }
     \vspace{-2mm}
    \begin{tabular}{|c| l l l | c|}
        \hline
        \multirow{2}{*}{Set} & \multirow{2}{*}{\# Objects} & Relative &  Clut. & \multirow{2}{*}{\# Samples}\\
         &  &  Clut. Width & Depth & \\
        \hline
        \textit{Trivial} & $< 3$ & $\ge 0$  & $\ge 1$ & 90798 \\
        \textit{Easy} &  $\ge$ 3 & $\ge 0$ &  $\ge 1$ & 186998 \\
        \textit{Medium} & $\ge$ 6 & $>$ 0.4 & $\ge$ 3 & 13616\\
        \textit{Hard} & $\ge$ 10 & $>$ 0.5 & $\ge$ 4 & 4884\\
        \hline 
    \end{tabular}
    \vspace{-2mm}
    \label{tab:splits}
\end{table}
\begin{table}[]
    \caption{
    Overview of datasets for visual manipulation relationship reasoning, highlighting D3GD as the most challenging
    }
    \vspace{-2mm}
    \centering
    \resizebox{0.9\linewidth}{!}{
    \begin{tabular}{|l| c c c c c|}
    \hline
       \multirow{2}{*}{Dataset} & Data & 
       Avg. & 
       Avg. 
       & Clutter & Clutter \\
       & Type & \#Rel & \#Objs & Depth & Width 
       \\
        \hline
        VMRD Train \cite{vrmn} & Real & 2.39 &	3.45 & 2.48 & 3.34 \\
        VMRD Test \cite{vrmn} & Real  & 2.35 & 3.40 &	2.44 &	3.28  \\
        REGRAD Train \cite{Regrad} & Synt  & 2.72 & 9.43 & 1.64 &	1.79 \\
        REGRAD Test \cite{Regrad}  & Synt  & 2.77 &	9.54 &	1.65 &	1.79 \\
        \hline
        D3GD Train & Synt  & 4.04 & 5.42 & 2.60 & 3.49  \\
        D3GD Test Easy  & Synt & 2.89 & 4.89 &	2.41 & 2.84 \\
        D3GD Test Medium  & Synt & 14.33 & 10.51 & 4.39 & 9.20  \\
        D3GD Test Hard & Synt &	30.40 & 16.05 & 4.46 & 15.05 \\
        D3GD Test Real  & Real & 2.58 & 3.70 & 1.09 & 2.23  \\
        \hline
    \end{tabular}
    }
\vspace{-6mm}
    \label{tab:data_comp}
\end{table}

We define our experimental testbed D3GD to get closer to real-world conditions where robots are challenged with grasping objects from large containers.
We build on MetaGraspNetV2 \cite{metagraspnetv2} which is composed of nearly 300K synthetic images and 2.3K real images depicting bin-picking scenes composed by up to 35 object instances sampled from 97 classes.   
The dataset comes with both 2D and 3D modalities as well as various types of grasp and labels.

For our use case, we consider only RGB images, 
the object bounding boxes, the grasping dependency graph, and, for the baseline models that use them, the instance segmentation annotations.
We organize the dataset into different difficulty levels based on three key metrics: \emph{number of objects}, \emph{clutter-depth}, and \emph{clutter-width}. 
\begin{figure}[t!]
    \centering
\includegraphics[width=0.75\linewidth]{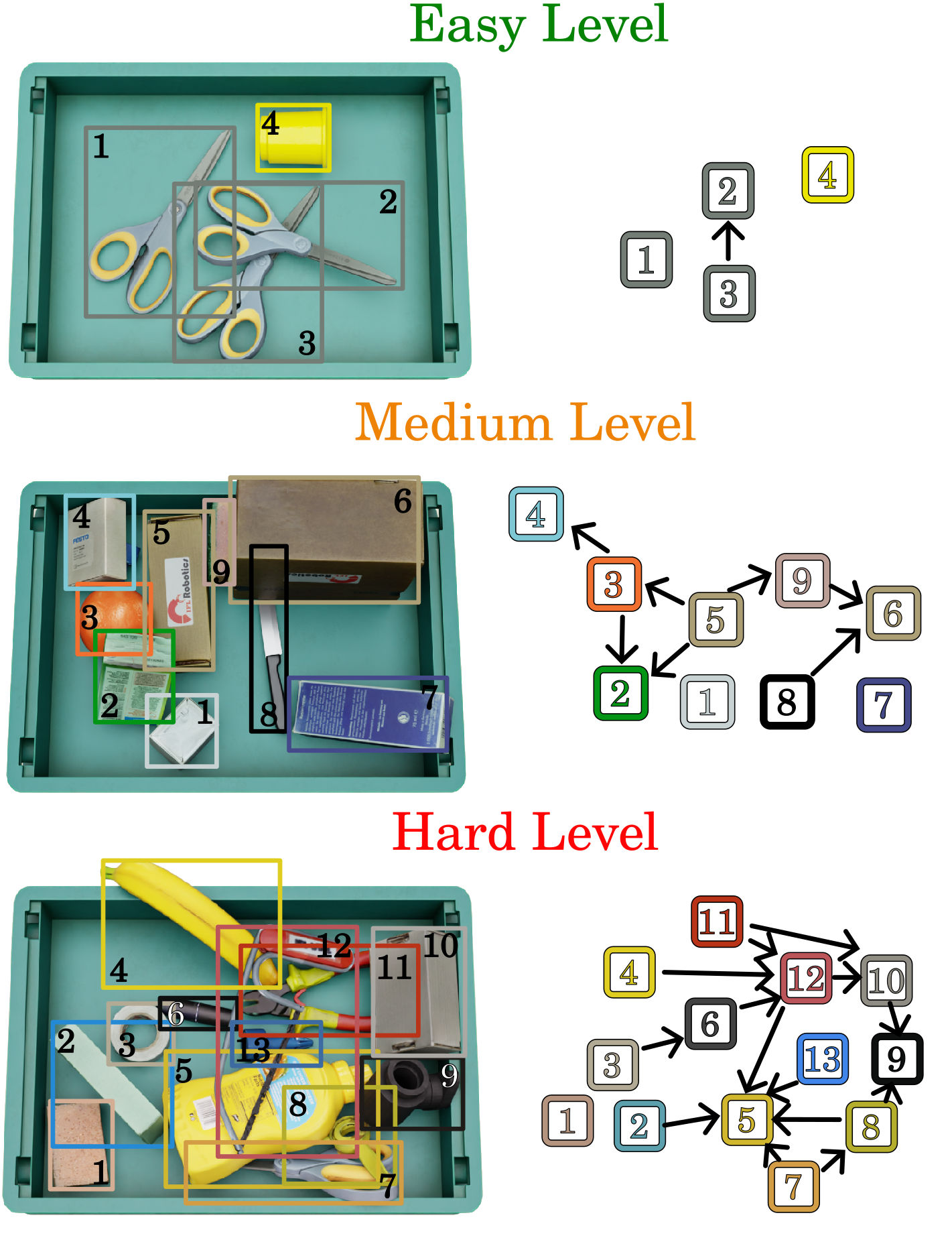}
    \vspace{-5mm}
    \caption{Data samples from the three main difficulty levels of D3GD. The objects in the bin on the left are identified by a number which is also used in the corresponding graph on the right. Here the arrows indicate a relationship dependency: A $\rightarrow$ B entails that A can't be moved without first moving B.
    In a downstream grasp task, the objects that should be touched first are those that don't have any outward relationships. For example, in the medium level, the green box (2) and brown box (6) can be safely grasped without causing object collapses that would alter the scene configuration.}
    \label{fig:test_samples}
    \vspace{-5mm}
\end{figure}

To consider the scene density and the challenges provided by object occlusion we measure 
how many objects (nodes) are part of a connected subgraph within the whole dependency graph. 
We measure the clutter-depth as the longest shortest path from a graph leaf (the first object to be picked) to the root (the last one). This means that we count the minimum number of objects that we need to remove before reaching an object at the bottom of the bin. As the dependency graph of a scene may have multiple leaves and roots, we annotate for each scene the longest of these paths which is a good indicator of how many layers of clutter the bin contains. 
Note that we are interpreting the dependency graph as a Directed Acyclic Graph (DAG) and we removed all the not-fitting scene samples (only $\sim$60) that, upon visual inspection, were clearly identified as errors produced during the synthetic dataset generation phase.
We also measure the clutter-width as the largest connected component: it yields a measure of how cohesive the clutter is, as seemingly complex scenes with many objects could become trivial when they are far apart and don't have a relationship between them.

Overall, our D3GD is organized in difficulty sets by choosing manual thresholds upon visual inspection of the samples. In Tab. \ref{tab:splits} 
we present the relative clutter width normalized by the number of objects in the scene.
For our study, we disregard the trivial set and focus on the easy, medium, and hard ones (see Fig. \ref{fig:test_samples}). 
We split the dataset into train, eval, and test scenes with an 80:10:10 proportion.  

Furthermore, we include in D3GD the real images of MetaGraspNetV2 as a cross-domain test set. Those samples (743 after filtering for missing and wrong annotations) are used to evaluate the generalization performance in a real-world scenario of models trained on synthetic data. 
 
To frame our D3GD in the existing literature we present a comparison with the VMRD \cite{vrmn} and REGRAD \cite{Regrad} datasets on the basis of the metrics used to define the difficulty levels of our testbed. 
Tab. \ref{tab:data_comp} shows how D3GD achieves complexities that weren't explored before. In particular, even if the number of objects of REGRAD \cite{Regrad} is relatively high, the complexity metrics are all fairly low. This can be attributed to its simulated tabletop setting, where cluttered piles easily fall apart leading to sparsely arranged objects.

In contrast, our setting is characterized by 
denser, more intricate graphs, driven by the physical constraints of the bin environment.

\section{Method}
In this section, we present our new visual relationship reasoning model, D3G, which is formalized as a transformer-based graph generation approach. It jointly predicts the graph nodes corresponding to objects in the scene, and the edges that encode their spatial relationships. 
Our end-to-end learning strategy avoids the drawbacks of two-stage pipelines \cite{vrmn, gvrmn_neuro, GruVMRN} and allows the model to determine which are the most relevant information from the scene to solve the task. 

We designed the architecture of D3G as a tailored integration of detection \cite{detr} and graph \cite{dwivedi2021generalization} transformers. The detection component extracts features through a combination of self and cross attention mechanisms, encoding them into a pre-defined number of object queries, set to exceed the maximum number of objects in the scene. 
These queries are pair-wise combined to represent object relationships.  
While incorporating an additional self-attention module to improve the relationship features via cross-talk could be beneficial, it becomes computationally prohibitive. 
Thus, we opted for a graph transformer that refines the relationship features iteratively, and we leveraged the fixed number of object queries to write its tensor operations into a dense format and obtain an efficient model. 

Finally, the object queries are matched to the detected objects: the optimal assignment procedure allows us to evaluate the node loss as in standard detection \cite{detr}, but we also extended its application to calculate the relationship loss. 
Specifically, the edge prediction matrix, initially generated through binary classification of all query pairs to determine relationship existence, is refined using a selection matrix defined by the optimal assignment. 

\begin{figure}[t!]
    \centering
    \includegraphics[width=0.95\linewidth]{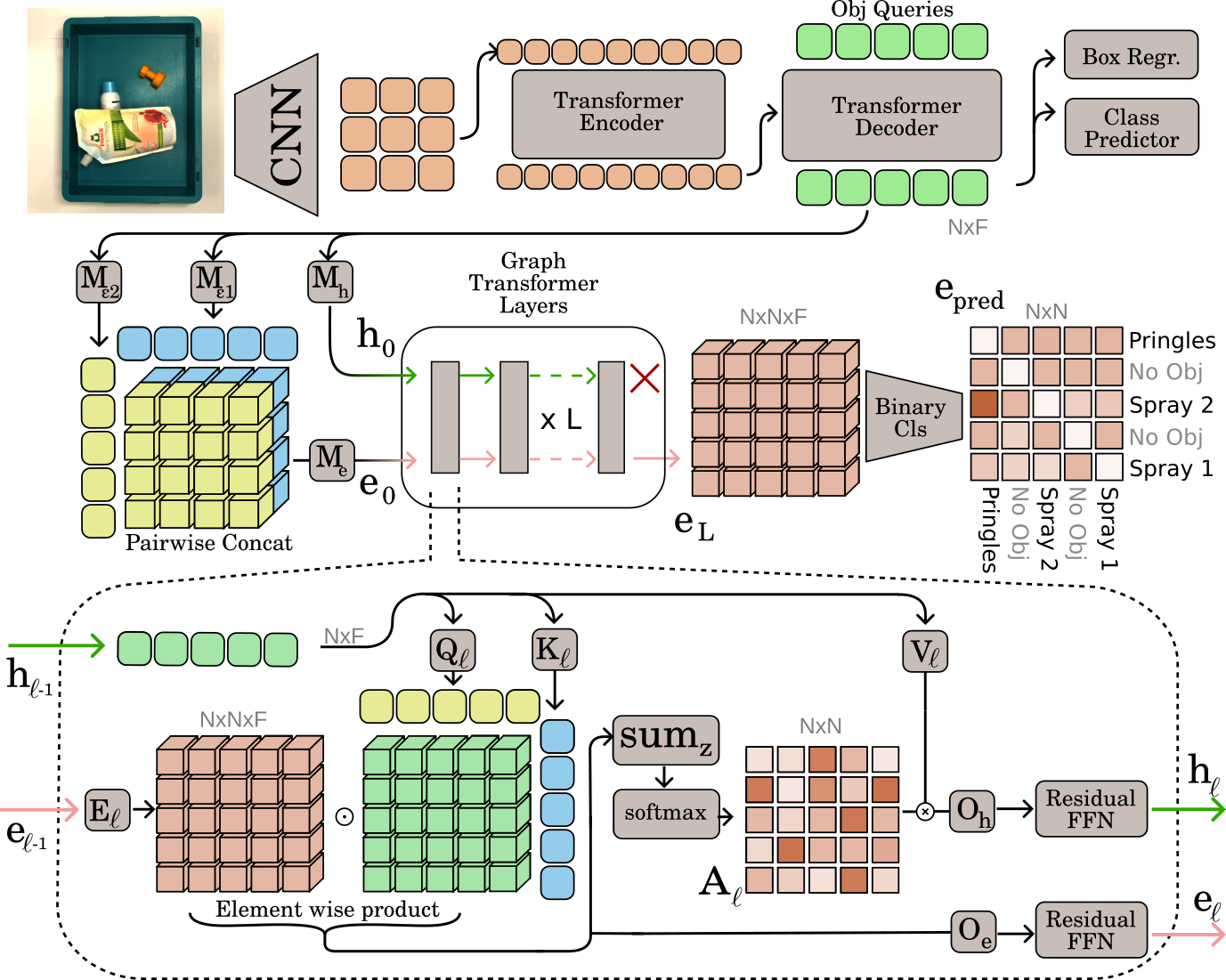}
    \caption{
    Illustration of our D3G. 
    \emph{The top row} describes the transformer architecture composed of encoder and decoder. The former takes as input scene features and provides as output updated representations which are fed to the decoder together with a fixed number of object queries. The query features obtained from the decoder are filtered and matched to the ground truth objects in the scene to detect them during training.  \emph{The middle row} shows how the same object query representations define the edge features via pairwise concatenation and mappings, before being refined by several graph transformer layers. Finally, a binary classifier predicts from the edge features whether two objects in the scene are connected by a spatial dependency relation.
    \emph{The bottom row} zooms inside a layer of the graph transformer assuming only one head for visual clarity. 
} \vspace{-5mm}
    \label{fig:model}
\end{figure}

Fig. \ref{fig:model} illustrates the proposed approach and helps to visualize the model flow explained in the following. 

\subsection{Detecting Objects as Graph Nodes} 
Given an input image of dimension $3 \times H_0 \times W_0 $, we perform object detection as a direct set prediction problem by following DETR \cite{detr}. A Convolution-based backbone extracts a compact feature representation of dimensionality $C\times H \times W$ (where $C=2048$, $H={H_0}/{32}$,  $W={W_0}/{32}$) which is rescaled via a $1\times 1$ convolution to reduce the channel dimensionality. The resulting $d\times H \times W$ tensor is interpreted as a list of vectors ($d\times HW$). 
These are fed to a transformer encoder \cite{attallneed} together with their spatial positional encodings and the obtained output is passed to the transformer decoder which also receives $N$ object queries (tensor of dim. $N\times F_0$, initially set to zero). The latter is transformed by the decoder in new embeddings that maintain the original dimension and are then processed by a feed-forward network into box coordinates and class labels. 
By utilizing self-attention and encoder-decoder attention mechanisms, the model globally reasons about all the objects at once, considering their pairwise relations while leveraging the entire image as context. 

\subsection{Predicting Object Relations as Graph Edges}
To elaborate the graph edges, we provide the $N$ embeddings produced by the decoder as input to a graph transformer inspired by \cite{dwivedi2021generalization}.
The node embeddings go through three different mappings operated by the matrices $M_h$, $M_{\epsilon1}$, $M_{\epsilon2}$ 
that work as adapters (each of dim. $F_0\times F$). We indicate their outputs respectively as $h_0$, $\epsilon_1$, $\epsilon_2$ (each of dim. $N\times F$). The last two are pairwise concatenated and then rescaled by halving the dimensionality via a further mapping $M_e$, to obtain $e_0 = M_e(\epsilon_1\oplus \epsilon_2)$. Both $h_0$ (dim. $N\times F$) and $e_0$ (dim. $N\times N \times F$) enter the first graph transformer layer and are progressively refined.

Each layer of our graph transformer operates similarly to the classical multi-head attention \cite{attallneed} but integrates edge features in its pipeline. A layer is characterized by an embedding of dimension $F$ and by $S$ heads. We conveniently define the dimension of the output of each head as $z=F/S$. We denote by $s=1,\ldots,S$ the index that runs over the $S$ heads, and by $l=1,\ldots,L$ the index over the $L$ layers. 

A layer takes as input the node features $h_{l-1}$ and edge features $e_{l-1}$ and elaborates them through three learned projection matrices (each of dim. $F \times z$), query $Q_{l,s}$, key $K_{l,s}$ and edge $E_{l,s}$, to produce a pseudo-attention matrix $\hat{A}_{l,s}$ (dim. $N \times N \times z$) whose $i,j$ elements are obtained as  
\begin{equation}
    \label{eq:edge_attention}
    \hat{A}_{l,s}^{ij} = \left(\frac{(Q_{l,s}\cdot h_{l-1}) \cdot {(K_{l,s} \cdot h_{l-1})}^\intercal}{\sqrt{z}}\right ) \cdot (E_{l,s} \cdot e_{l-1}^{ij})~.
\end{equation}
By summing over the third feature dimension ($z$) and applying a column-wise softmax operation we get the attention matrix in standard format (dim. $N \times N$):\vspace{-1.5mm}
\begin{equation}
    \label{eq:softmax}
    A_{l,s} = \text{softmax}(\text{sum}_z(\hat{A}_{l,s}))~. \vspace{-1.5mm}
\end{equation}
We remark that operations analogous to those in Eq. (\ref{eq:edge_attention}) and (\ref{eq:softmax}) are managed in \cite{dwivedi2021generalization} by costly message passing layers. Instead, we are dealing with dense matrix operations performed via matrix multiplications which are easily parallelizable. This is possible because in our formulation we have a fixed set of queries as nodes in a fully connected graph.

Finally, the layer output is composed of the updated node and edge features obtained respectively as:\vspace{-1mm}
\begin{align}\label{eq:hhat}
    h_{l} &=  (\Vert_{s=1}^S A_{l,s} \cdot h_{l-1} \cdot V_{l,s})~ O_h^l  \\
    e_{l} &=  (\Vert_{s=1}^S \hat{A}_{l,s})~ O_e^l~. \vspace{-1mm}
\end{align}
Here $V_{l,s}$ is the learned value matrix (dim. $F \times z$), $O_h^l,O_e^l$ are projection matrices (dim. $F \times F$), and $\Vert$ denotes concatenation which is executed over the $S$ heads. The outputs are then passed to a Feed Forward Network (FFN) preceded and succeeded by residual connections and normalization layers. Due to our particular use case, at the last layer we keep only the edge output $e_{l=L}$ and disregard the node output $h_{l=L}$. 
The edge features are fed to an MLP that maps each of them 
to a logit indicating whether that edge exists, i.e. the pair of nodes connected by that edge have a dependency relationship. The obtained matrix $e_{pred}$ has dimension $N \times N$ and represents the adjacency matrix of a graph with $N$ nodes. 

\begin{figure}[t!]
    \centering
    \includegraphics[width=0.9\linewidth]{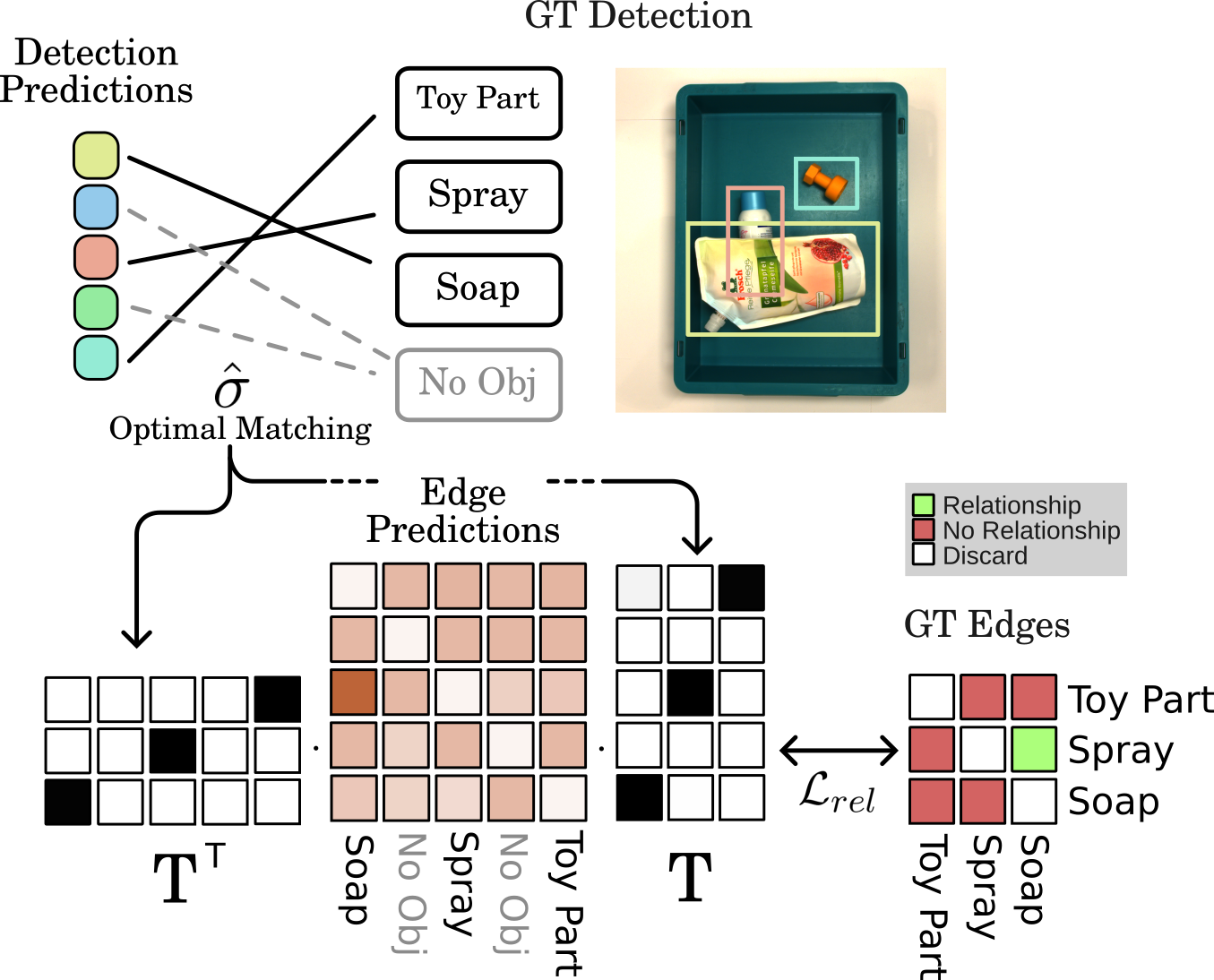}
        \vspace{-3mm}
    \caption{ 
    Relationship loss calculation procedure. 
    The edge prediction matrix $e_{pred}$ contains a score for each pair of object queries. Only some of the N=5 queries match the real P=3 objects. 
    The optimal bipartite matching $\hat{\sigma}=[5,3,1,2,4]$ 
    identifies the correspondence between the predictions $\hat{y}=[3,0,2,0,1]$ and the ground truth objects $y=[1,2,3,0,0]$, which are in order Toy Part, Spray, and Soap.
    Here $T$ has dimensions $5\times3$ and has value 1 in the cells that identify a match: e.g. cell (5,1) will contain a 1 because the object with ID 1 has been matched to the 5-th prediction. The matrix obtained by multiplying $T^\top \cdot e_{pred} \cdot T$  
    is directly comparable with the GT matrix via a simple cross-entropy loss.}
    \vspace{-5mm}
    \label{fig:enter-label}
\end{figure}

\subsection{Loss Function}
To train our D3G we evaluate jointly its object detection and object relationship predictions,  and progressively optimize them.
The loss function measures both aspects and must accommodate the model's output on a set of 
$N$ bounding boxes, which exceeds the number of objects $P<N$ in an image. Thus, we need first of all to search for the optimal match between them. 

We define as $y$ and $\hat{y}$ the set of ground truth and predicted objects. They are both of dimension $N$, with the former containing $P-N$ positions padded with  $\emptyset$ (no object). More precisely, it holds $y_i=(c_i,b_i)$, where the class label $c_i$ may be $\emptyset$, and the ground truth box $b_i \in [0,1]^4$ is composed of the center coordinates as well as its height and width relative to the image size. 
By following \cite{detr}, we use the Hungarian algorithm and search for the optimal bipartite matching $\hat \sigma$ from all possible permutations of the $N$  elements $\sigma \in \mathfrak{G}_N$: \vspace{-1mm} 
\begin{equation}
    \hat \sigma = {\arg \min}_{\sigma \in \mathfrak{G}_N} \sum_i^N\mathcal{L}_{match}(y_i,\hat{y}_{\sigma(i)})~,
    \vspace{-3mm}
\end{equation}
\begin{align}
\text{with}\quad\mathcal{L}_{match}(y_i,\hat{y}_{\sigma(i)})= &-\mathds{1}_{\{c_i \neq \emptyset\}}\hat{p}_{\sigma(i)}(c_i) \nonumber \\
&+\mathds{1}_{\{c_i \neq \emptyset\}}\mathcal{L}_{\text{box}}(b_i,\hat{b}_{\sigma(i)})~.
\end{align}
Here $\hat{p}_{\sigma(i)}(c_i)$ is the probability of class $c_i$ for the prediction with index $\sigma(i)$, and $\hat{b}_{\sigma(i)}$ is the related bounding box. The loss component $\mathcal{L}_{\text{box}}$ is a linear combination of a $\ell_1$ loss and a generalized scale-invariant $\text{IoU}$ loss \cite{giou}. 
Once the matching is obtained, we leverage the optimal assignment $\hat\sigma$ to calculate the \emph{node loss} that measures the accuracy in detecting the objects that define the nodes of the dependency graph, and the \emph{relationship loss} that evaluates the accuracy in the edge prediction. 
The first is the sum of a negative log-likelihood for class prediction and box loss \cite{detr}:\vspace{-1mm} 
\begin{equation}
\mathcal{L}_{node}(y,\hat{y})=  \sum_{i=1}^N \left[-\log \hat{p}_{\hat{\sigma}(i)}(c_i) +\mathds{1}_{\{c_i \neq \emptyset\}}\mathcal{L}_{\text{box}}(b_i,\hat{b}_{\hat{\sigma}(i)})\right],
\end{equation}
it differs from $\mathcal{L}_{match}$  by the use of the log-probability in the first term and by calculating a cost for all the classes, including the assignment to a void object with class $\emptyset$. 

In our preliminary experiments on edge relationships we observed that simply reordering the ground truth to align with our predictions was insufficient. The presence of numerous empty relationships, stemming from non-object boxes, greatly outnumbered true relationships and caused the models to converge very slowly. To address this, we developed a relationship loss function that simultaneously filters out and reorders the relationships.
We start by introducing the rectangular \emph{selection matrix} $T$ of dimensions $N \times P$ whose elements are defined as $T_{ij} = \mathds{1}_{\{i = \hat\sigma(j)\}}$, 
where $i$ and $j$ run respectively over rows and columns (see Fig. \ref{fig:enter-label}). We use it to focus only on the identified ground truth objects that represent the nodes of the dependency graph and evaluate the related edges via the 
predicted adjacency matrix: \vspace{-1mm}
\begin{equation}
    \hat{G} = T^\intercal\cdot e_{pred} \cdot T~.
    \label{eq:G} 
\end{equation}
Finally, we calculate the relationship loss using a simple binary cross entropy that indicates whether we correctly identify the presence or absence of an edge:\vspace{-1mm}
\begin{equation}
    \mathcal{L}_{rel} = \sum_i^P \sum_j^P \mathcal{L}_{bce}(G_{ij},\hat{G}_{ij})~,\vspace{-1mm}
\end{equation}
where the binary ground truth matrix $G$ represents the object relationships and their direction ($G_{ij}\neq G_{ji}$). 
Overall, the total loss formulation for our D3G is 
$\mathcal{L}_{tot} = \mathcal{L}_{node} + \mathcal{L}_{rel}$. 

\section{Metrics}
\label{metrics}
The task of visual manipulation relationship reasoning is usually evaluated by the \textbf{mAP} detection score, that is the mean of the Average Precision over all the object categories at a given Intersection over Union. This metric is calculated by sweeping over all the bounding boxes assigned to a certain class sorted by probability (i.e. only the relative score matters). 
Since the absolute probability value is highly sensitive to the calibration process, disregarding it allows for a more objective comparison among different approaches. 

The Object triplet Recall (\textbf{OR}) and Precision (\textbf{OP}) are computed on 
object pairs, considering the relational triple $(obj_i, r_{ij}, obj_j)$ true when all elements are correctly predicted. 
Specifically, the edge between two objects may have three possible labels $r: (in, out, no\_rel)$. OR and PR are the average recall and precision of all three kinds of manipulation relationships. 
These metrics depend on an object prediction score threshold, used to determine whether a bounding box contains an object at $\text{IoU}>0.5$.
This threshold is typically tuned on the validation set which might not be available or be very dataset-specific, leading to generalization challenges.

\newcommand{\mapcoco}{$\underset{\text{coco}}{\text{mAP}}$}

\begin{table*}[!ht]
\centering
\caption{Results on the D3GD synthetic and real sets. Best results per column highlighted in bold and second best underlined.}
    \vspace{-2mm}
\resizebox{0.95\textwidth}{!}{
\begin{tabular}{| c | c | l p{4mm} p{4mm} p{5mm} | l p{4mm} p{4mm} p{5mm} | l p{4mm} p{4mm} p{5mm} | l p{4mm} p{4mm} p{5mm} |}
\hline
\multirow{2}{*}{Model} & \multirow{ 2}{*}{Detector Base} & \multicolumn{4}{ c |}{Synth Easy} & \multicolumn{4}{ c |}{Synth Medium} & \multicolumn{4}{ c |}{Synth Hard} & \multicolumn{4}{ c |}{Synth-to-Real} \\
& & \mapcoco \hspace{-3mm} & $\text{AP}_{\text{rel}} $ & OR & OP & \mapcoco \hspace{-3mm} & $\text{AP}_{\text{rel}}$ & OR & OP & \mapcoco \hspace{-3mm} & $\text{AP}_{\text{rel}}$ & OR & OP & \mapcoco \hspace{-3mm} & $\text{AP}_{\text{rel}}$ & OR & OP  \\  \hline
VMRN~\cite{vrmn} & \multirow{3}{*}{Mask-RCNN~\cite{mask-rcnn}} & \underline{86.96} & 73.29 & \underline{88.76} & 86.12 & \underline{75.20} & 54.97 & \underline{70.22} & \underline{71.86} & \textbf{67.36} & 42.49 & \underline{62.26} & 71.58 & 45.04 & 13.36 & \underline{49.70} & 10.20 \\ 
GVMRN~\cite{gvrmn_neuro} &  & 86.39 & 57.22 & 83.27 & 80.58 & 74.03 & 34.89 & 62.63 & 63.30 & 66.05 & 21.54 & 54.34 & 60.50 & 43.76 & 7.62 & 49.39 & 10.30 \\
GGNN~\cite{GruVMRN} &  & 86.31 & 70.51 & 87.41 & 84.15 & 74.19 & 51.43 & 68.04 & 69.39 & 66.02 & 37.52 & 59.11 & 68.58 & 44.50 & 10.55 & 39.03 & 7.11 \\ 
\hline
EGTR~\cite{im2024egtr} & Def. DETR~\cite{zhu2020deformable} & 84.23 & 52.47 & 79.28 & 78.20 & \textbf{75.37} & 33.13 & 61.33 & 70.11 & \underline{66.66} & 22.72 & 48.07 & \underline{73.38} & \textbf{56.36} & 15.11 & \textbf{56.25} & \underline{48.88}  \\ 
\hline
\multirow{ 2}{*}{D3G (ours)} & DETR \cite{detr} & 83.22 & \underline{76.27} & \textbf{89.57} & \underline{88.64} & 70.92 & \underline{58.65} & \textbf{72.86} & 71.58 & 60.25 & \underline{43.99} & \textbf{63.71} & 57.85 &  55.93 & \textbf{24.44} & 41.11 & \textbf{51.36} \\ 
 & Def. DETR~\cite{zhu2020deformable} & \textbf{87.68} & \textbf{76.96} & 87.42 & \textbf{90.66} & 74.84 & \textbf{60.81} & 66.91 & \textbf{77.98} & 65.92 & \textbf{46.45} & 51.69 & \textbf{85.18} & \underline{56.14} & \underline{22.38} & 33.70 & 38.90 \\
 \hline
  \end{tabular}
  }
  \vspace{-2mm}
  \label{tab:synthres}
\end{table*}

To address this limitation, we propose to use the \emph{Average Precision of Relationships}  (\textbf{AP$_{\text{\textbf{rel}}}$}) that extends the AP used for detection to work on the prediction of object triplets. A  triplet is identified as True Positive if both the object bounding boxes have $\text{IoU}>0.5$ with a previously undetected ground-truth annotation, and all three labels match their corresponding ground-truth. Any other case is considered a False Positive. Using this definition, the Precision-Recall curve is calculated by varying the detection confidence threshold, and the AP$_{\text{rel}}$ is the area under such curve. Thus, it uses the same criteria as the OR and OP, but rather than taking a single point on the curve given an arbitrary threshold, it sweeps on all possible precision-recall tradeoffs yielding a unique threshold-free score.
On Open Images \cite{kuznetsova2020open} for visual relationships detection the $\text{AP}_\text{rel}$ provided very pessimistic results due to the extensive variety of scenes and multiple valid semantic relations that may exist between the same pairs of objects, making comprehensive labeling extremely challenging. 
This is not the case for our task and data where the object relationships can be exhaustively annotated.

\section{Experiments}

\subsection{Reference Baselines}
We compare our D3G with three previous approaches all composed of separate object detection and relationship reasoning stages. 
\textbf{VMRN} \cite{vrmn} originally exploited a Single Shot Detector (SSD~\cite{Liu2015SSDSS}).  The obtained bounding boxes and related convolutional features are provided as input to an Object Pairing Pooling Layer to classify manipulation relationships. Both \textbf{GVMRN} ~\cite{gvrmn_neuro} and \textbf{GGNN}~\cite{GruVMRN} build on Faster-RCNN~\cite{faster-rcnn} including an object pair pooling to provide paired features to a graph neural network. 
For a fair comparison, we follow the best practice proposed in \cite{metagraspnetv2}, considering Mask-RCNN with ResNet50 backbone as detector for all these three approaches and adding the respective network heads. We highlight that Mask-RCNN exploits instance segmentation as an auxiliary objective and needs the related data annotations. 

We also add a fourth baseline from the scene graph generation literature.
The recent \textbf{EGTR} \cite{im2024egtr} presented a lightweight object relation extractor from the decoder of a deformable DETR detector \cite{zhu2020deformable}. 
A multi-task learning strategy allows to predict jointly the object location and category as well as the relationship label.  
As we will discuss in the result subsection, the graph transformer layer exploited by our D3G allows to better manage complex dependency relations for grasp planning. 
We evaluate D3G when built on DETR and on deformable DETR from a ResNet50 backbone.  

\subsection{Training Procedure Details and Experimental Setup}
We train all the considered methods by first focusing on the detection modules and reaching convergence.
This process takes around 12 epochs for Mask-RCNN and 18 epochs for the DETR based architectures. The relational reasoning components are then trained while further fine-tuning detection with a lower learning rate for 6 epochs.
For all phases we use AdamW optimizer \cite{loshchilov2018decoupled}. We tune all the hyperparameters by monitoring the mAP and the $\text{AP}_{\text{rel}}$ on the validation set. Finally, we report all results on the test splits when using the chosen hyperparameters and training on the combined set of training and validation data. 
We use random flip, brightness, contrast, and saturation data augmentations.
We conduct all experiments on a single GPU NVIDIA A100(40GB). 

\begin{figure*}[t!]
    \centering
    \includegraphics[width=0.85\linewidth]{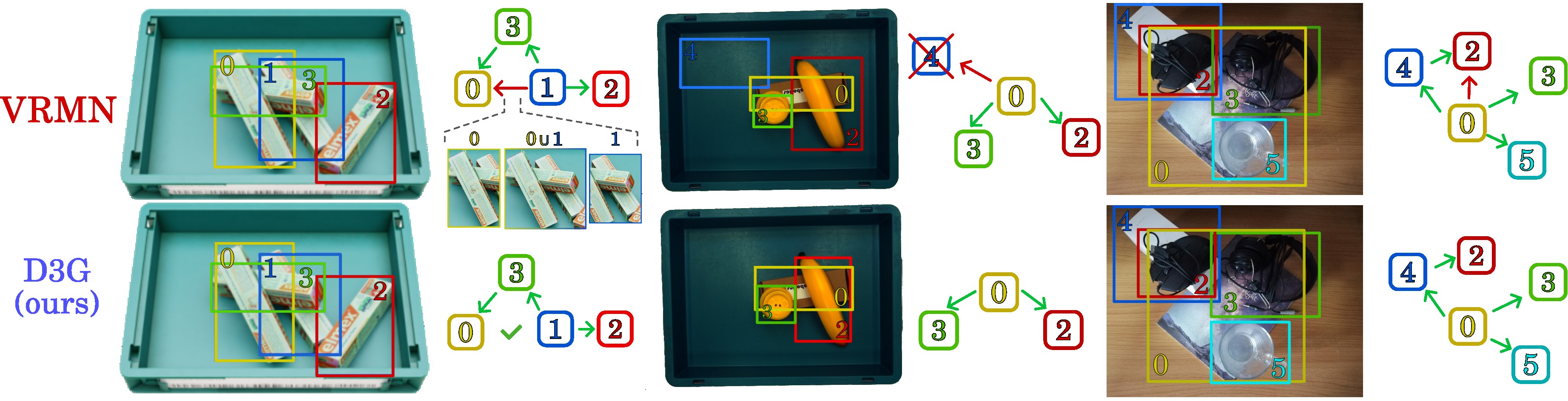}
    \vspace{-4mm}
    \caption{Qualitative example of VMRN vs \ours. 
    Left: DG3D synthetic test set. Here the object localization is almost identical and correct in the two cases but VMRN shows a wrong relationship between objects 0 and 2. The features obtained from the union area of the two bounding boxes include an intruding object causing confusion.
    Middle: DG3D synth-to-real test set. Our model shows good synth-to-real transfer while VMRN suffers from the domain shift and detects a false positive.
    Right: VMRD test set. Both models perform well on the real data but VMRN predicts a wrong relationship.
    }
    \label{fig:qualitative}
    \vspace{-5mm}
\end{figure*}

\subsection {Results on D3GD}
In Tab. \ref{tab:synthres} we report results obtained on our D3GD. 
By focusing on the mAP columns for the Synth cases we can notice that the detection performances of the Mask-RCNN-based methods show minimal differences among each other, with VMRN generally obtaining the best result. It is well-documented that the DETR architecture has some limitations that prevent it from achieving optimal performance compared to well-tuned two-stage detectors. 
These issues are mitigated by the introduction of the deformable attention \cite{zhu2020deformable} which shows an improved ability to focus on the most relevant local areas of the image: the beneficial effects appear in the mAP scores of EGTR as well as our D3G. 
In terms of AP$_\text{rel}$, OR, and OP, VMRN overcomes the other Mask-RCNN-based approaches as well as EGTR, showing a particularly evident advantage in the hard split. Yet, for all those metrics, D3G consistently shows the best results.
This performance improvement can be attributed to the formalization of the relationship features of D3G. The competitor two-stage methods rely on cropping the feature maps at the bounding box locations of the two objects and concatenating them: in Fig. \ref{fig:qualitative} we show how this can lead to wrong relationships when the crops are ambiguous due to high clutter. Our D3G can extract meaningful features thanks to the transformer-decoder cross-attention modules solving such cases. 

In the Synth-to-Real setting, there is an evident performance drop with respect to the synthetic case due to the domain shift between the training and test data. 
Nevertheless, the DETR-based methods present better detection and relationship reasoning abilities than the Mask-RCNN-based competitors, with D3G as the best in AP$_\text{rel}$ and second best in mAP. 
We underline that the lack of a validation set in this Synth-to-Real scenario makes it impossible to search for the optimal detection score threshold for the definition of OR and OP. As a consequence,  the values for these metrics can exhibit considerable variations which make it difficult to draw conclusions from them. 

\subsection{Analysis of Design Choices} 
In designing our D3G, we made specific choices regarding how the representations of two nodes (objects) are combined to define an edge (relationship) and the composition of the layers in the graph transformer module. Here we discuss these choices, showing the impact of alternative options. 

\begin{table}[t]
    \centering
        \caption{
        D3G $\text{AP}_{\text{rel}}$ 
        when changing the edge initialization technique.}
            \vspace{-2mm}
        \label{tab:edge_init}
    \begin{tabular}{|l | c c c c|}
    \hline
         {$e_0$ init. strategy} & {S.Easy} & {S.Medium} & {S.Hard} & {Real} \\
         \hline
         Hadamard product & 76.20 & 58.93 & \underline{44.23} & \underline{24.09} \\
         Sum & 75.96 & \underline{59.51} & 43.61 & 21.45 	 \\
         Difference & \textbf{76.42} & \textbf{59.81} & \textbf{44.41} & 23.59 \\
         \hline
         Pairwise Concat. & \underline{76.27} & 58.65 & 43.99 & \textbf{24.44} \\
\hline
    \end{tabular}
\vspace{-3mm}
    
\end{table}
\begin{table}[t]
    \centering
    \caption{
    D3G $\text{AP}_{\text{rel}}$
    at different head sizes $S$ and num. of layers $L$.}
    \label{tab:transformer_dims}
    \vspace{-2mm}
    \resizebox{0.9\linewidth}{!}{
    \begin{tabular}{| l@{~~~~}l | c | c@{~~~~} c@{~~~~} c@{~~~~} c|}
    \hline
         \multicolumn{2}{|c|}{Graph Transf.}
         & \# Params & {S.Easy} & {S.Medium} & {S.Hard} & {Real} \\
         \hline
\multirow{3}{*}{S=2}
        & L=1 & 0.921M & 76.26 & 58.08 & 44.30 &  \textbf{25.32} \\
        & L=2 & 1.84M & \textbf{76.77}  & 58.65 & 43.99 & 24.44 \\
        & L=4 & 3.68M &  76.06  & 59.32   & 44.25  & 25.37   \\
        \hline
\multirow{3}{*}{S=4}       
        & L=1 & 0.921M &    76.40   & \textbf{60.00}  & 45.27 & 23.18  \\
        & L=2 & 1.84M & 76.10    & 54.84  & 44.17  & 24.66   \\
        & L=4 & 3.68M &73.38 & 57.68 & \textbf{50.31} & 24.32 \\
         \hline
    \end{tabular}
    }
    \vspace{-5mm}
\end{table}

To define $e_0$, D3G exploits the pairwise concatenation of $\epsilon_1$ and $\epsilon_2$. We tested three other strategies: Hadamard product, Sum, and Difference. 
The results, presented in Tab. \ref{tab:edge_init}, show minimal variability. 
The best are pairwise concatenation and pairwise differences, we choose the former as a good tradeoff between synthetic and real performance.

The graph transformer at the heart of our D3G can come in different sizes depending on the number of heads $S$ in the multi-head self-attention of each layer, and on the number of layers $L$. We adopted $S=2$, $L=2$, but we can evaluate a smaller ($L=1$) and larger ($L=4$) version. 
We also test the effect of increasing the number of heads to $S=4$. 
The results in Tab. \ref{tab:transformer_dims} indicate that on the easy set the results remain almost stable except when the number of heads and layers are too high ($S=4,L=4$) which leads to overfitting. On the medium and hard sets, increasing the number of heads can be beneficial but may require a significant increase in model parameters (e.g. $S=4,L=4$ for the hard case). 
Finally, results on real data suggest preferring fewer heads ($S=2$), and indicate that reducing the number of layers may provide further advantage ($L=1$ vs $L=2$). 

\begin{table}[t!]
    \caption{Analysis of D3G and its competitors on the VMRD \cite{vrmn} dataset.}
        \vspace{-2mm}
    \centering
    \resizebox{0.9\columnwidth}{!}{
    \begin{tabular}{|c | c c c c c|}
         \hline
         Model & $\underset{\text{voc}}{\text{mAP}}$ & OR & OP & IA & Time(ms) \\
         \hline
         VMRN \cite{vrmn} & 95.40 & 85.40 & 85.50 & 65.80 & \textbf{10}\\
         GVMRN \cite{gvrmn_neuro} & 95.40 & 87.4 0& 87.90 & 69.30 & 67 \\
         GGNN \cite{GruVMRN} & 96.40 & \textbf{90.09} & 88.01 & 75.33 & 130 \\
         \hline
         D3G  & \underline{96.50} & 88.88 & \textbf{91.12} & \textbf{78.33} & \underline{49} \\
         D3G-Deformable & \textbf{96.66} & \underline{89.64} & \underline{90.00} & \underline{77.55} & 87 \\
         \hline
    \end{tabular}
    }
\vspace{-6mm}
    \label{tab:vmrd_results}
\end{table}

\subsection{Results on VMRD}
We extended the experimental analysis on the standard VMRD dataset and report the results in Tab. \ref{tab:vmrd_results}, by considering the PascalVOC mAP as done in previous works.  
OR and OP are the same metrics presented in Sec.~\ref{metrics}, while IA (Image Accuracy) measures the fraction of dependency graphs that are fully reconstructed. D3G reports competitive values in mAP and OR, and attains the highest scores in OP and IA. Notably, the substantial improvement in IA highlights our method’s strong ability to accurately reconstruct complete dependency graphs.
In the last column of Tab. \ref{tab:vmrd_results} we also present 
the inference time of our method using a 2080Ti GPU, against competitors as claimed in their work: D3G exhibits competitive speed in both its proposed versions.
\section{Conclusions}
In this work, we advance the task of visual manipulation relationship reasoning by introducing a new testbed D3GD for more complex scenes than previously studied, and presenting a new model D3G. Unlike standard cascade approaches that separate object detection and relationship reasoning, our D3G end-to-end model leverages a transformer-based detector and graph layers to jointly predict object locations, class labels, and relationships. We also advocate for threshold-free metrics, such as Average Precision of Relationships, to ensure reliable evaluation, especially in real-world scenarios lacking validation sets for threshold tuning.
We publicly release all the setting details of our testbed and experimental analysis, together with the code suite to foster reproducibility. 

While the performance of our model on the proposed benchmark is remarkable there is still space for improvement. We plan to further improve the model's efficiency via subquadratic attention strategies and extend the task by incorporating multimodal signals such as 3D cues and language.





\smallskip
\textbf{Acknowledgements.}
This work was carried out within the FAIR - Future Artificial Intelligence Research and received funding from the European Union Next-GenerationEU (PIANO NAZIONALE DI RIPRESA E RESILIENZA (PNRR) – MISSIONE 4 COMPONENTE 2, INVESTIMENTO 1.3 – D.D. 1555 11/10/2022, PE00000013.

\bibliographystyle{IEEEtran}
\bibliography{bibliography}

\end{document}